\documentclass[12pt]{article}

\usepackage{arabtex}

\usepackage[T1]{fontenc}
\usepackage{inputenc}

\usepackage{coling2018}

\usepackage{url}
\usepackage[para,online,flushleft]{threeparttable}
\usepackage{booktabs}
\usepackage{float}
\usepackage{hhline}
\usepackage{graphicx,subcaption}
\usepackage{smartdiagram}
\usepackage{pgf-pie}
\usepackage{multicol}
\usepackage{enumitem}

\usepackage{caption} 

\usepackage{colortbl}

\usepackage[stable]{footmisc}

\usepackage{cp1256,utf8}
\setcode{utf8}
\usepackage{pgfplots}
\pgfplotsset{compat=1.18}

\date{}

\begin{document}

\title{I can tell whether you are a Native Hawl\^eri Speaker!\\How ANN, CNN, and RNN perform in NLI-Native Language Identification}

\author{
	\begin{tabular}[t]{c}
		Hardi Garari and Hossein Hassani\\
		\textnormal{University of Kurdistan Hewl\^er}\\
		\textnormal{Kurdistan Region - Iraq}\\
		{\tt {\{hardi.jawhar, hosseinh}\}@ukh.edu.krd}
	\end{tabular}
}

\maketitle

\begin{abstract}
Native Language Identification (NLI) is a task in Natural Language Processing (NLP) that typically determines the native language of an author through their writing or a speaker through their speaking. It has various applications in different areas, such as forensic linguistics and general linguistics studies. Although considerable research has been conducted on NLI regarding two different languages, such as English and German, the literature indicates a significant gap regarding NLI for dialects and subdialects. The gap becomes wider in less-resourced languages such as Kurdish. This research focuses on NLI within the context of a subdialect of Sorani (Central) Kurdish. It aims to investigate the NLI for Hewl\^eri, a subdialect spoken in Hewl\^er (Erbil), the Capital of the Kurdistan Region of Iraq. We collected about 24 hours of speech by recording interviews with 40 native or non-native Hewl\^eri speakers, 17 female and 23 male. We created three Neural Network-based models: Artificial Neural Network (ANN), Convolutional Neural Network (CNN), and Recurrent Neural Network (RNN), which were evaluated through 66 experiments, covering various time-frames from 1 to 60 seconds, undersampling, oversampling and cross-validation. The RNN model showed the highest accuracy of 95.92\% for 5-second audio segmentation, using an 80:10:10 data splitting scheme. The created dataset is the first speech dataset for NLI on the Hewl\^eri subdialect in Sorani Kurdish dialect, which can be of benefit to various research areas.

\end{abstract}

\section{Introduction}
Native Language Identification (NLI) is a task in Natural Language Processing (NLP) to automatically identify native speakers from non-native ones based on their foreign language production  \cite{malmasi2017report,Vajjala2018TheRO}. NLI essentially focuses on determining a speaker's or writer's native language based on their use of a second language, using various NLP techniques to analyze linguistic patterns, grammar, and other language features \cite{inbook}. The concept behind NLI is based on assumptions that the mother tongue influences Second Language Acquisition (SLA) and its production \cite{Malmasi2016}. NLI seeks to reproduce the human ability in recognizing nativity in computational ways \cite{Malmasi2016}. In fact, humans identify languages through an inherent auditory process involving various perceptual or psychoacoustic cues. As a result, the perceptual signals that listeners rely on often serve as a key source of inspiration in developing models to automatically identify spoken languages \cite{li2013spoken}.

NLI has various applications, for example, law enforcement authorities \cite{malmasi2017multilingual}, immigration offices to validate the claims of individuals about their origins and backgrounds. Subdialect identification is more challenging than Language identification because of the fine-grained linguistic variations between dialects of the same language \cite{khurana17_interspeech}. Both statistical machine learning (\cite{malmasi2017multilingual}) and NN-based approaches (\cite{isam2024you}) have been used in subdialectal classification. We extend this concept to apply it at the subdialect level in the Sorani (Central Kurdish). We develop three models based on Neural Networks (NN) to identify native Hewl\^eri speakers from non-native ones. The Sorani that is spoken in the Kurdistan Region of Iraq (KRI) is divided into six distinct subdialects: Hewl\^eri, Garmiani, Karkuki, Khoshnawi, Pishdari, and Sulaimani \cite{isam2024you}. Hewl\^eri has bee named after the Hewl\^er (Erbil), the KRI capital \cite{khorshid2018}.

The rest of this paper is organized as follows. Section 2 addresses the related work, Section 3 explains the method, Section 4 presents and discusses the experiments and results, and Section 5 concludes the paper.

\section{Related work}

While NLI started to emerge in the 2000s, it finds its roots in the works such as the one by \newcite{flege1984detection}, which investigated how American English listeners identified French accents. Through controlled experiments, the researcher demonstrated that phonetic and suprasegmental cues were effective in distinguishing accented speech, achieving 89\% accuracy. This was foundational work that bridged human perceptual capabilities with computational speech recognition systems.

\newcite{tetreault-etal-2013-report} reports on the first shared task on NLI, where `'a total of 29 teams from around the world
competed across three different sub-tasks'` on several languages.

\newcite{malmasi2014arabic} focused on less-studied languages such as  Arabic. They used the Arabic Learner Corpus (ALC) and employed supervised multi-class classification approaches, achieving 41\% accuracy. Despite having limited data and Arabic's linguistic complexity, their study demonstrated the capability the usage of machine learning for NLI.

\newcite{malmasi2014chinese} also explored NLI for Chinese learners using the Chinese Learner Corpus (CLC) by leveraging syntactic features such as POS n-grams, function words, and context-free grammar rules. Their model achieved 70.61\% accuracy. 


\newcite{rajpal2016native} showed how non-native English speakers' linguistic patterns are influenced by their native languages through utilizing features such as Mel Frequency Cepstral Coefficients (MFCC), Convolutional Restricted Boltzmann Machines (ConvRBM), and prosodic components such as phrase dynamics taken from speech signals. They used Gaussian Mixture Models (GMM) and achieved a Weighted Average Recall (WA) of 40.2\%. The findings highlighted the influence of spectral dynamics and prosodic cues in improving classification performance.

\newcite{isam2024you} explored three deep learning models: Artificial Neural Networks (ANN), Convolutional Neural Networks (CNN), and Recurrent Neural Networks with Long Short-Term Memory (RNN-LSTM) to classify Kurdish Sorani subdialects (Hewl\^eri, Garmiani, Karkuki, Khoshnawi, Pishdari, and Sulaimani) using a speech dataset.  Their collected dataset, \textit{Sorani Nas}, consists of more than 29 hours of audio recorded by native speakers. The RNN-LSTM model have the highest accuracy at 96\%, followed by the CNN model with 93\%. 

In summary, while probabilistic approaches, such as GMM, are still popular in NLI, there is a noticeable trend toward NN-based methods. We also use NN-based in this research.
 
\section{Method}

\begin{figure} [H]
    \centering
    \includegraphics[width=1\linewidth]{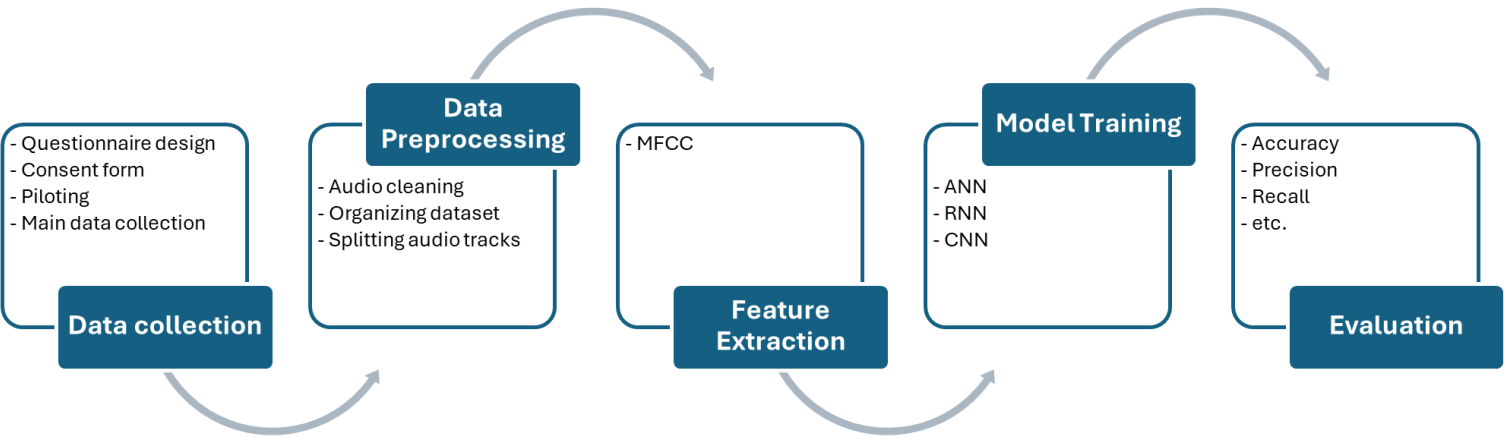}
    \caption{An overview of the method}
    \label{fig:rmd}
\end{figure}

This section presents the research method. Figure \ref{fig:rmd} provides an overview of the approach. A brief explanation for each step follows. 

We record interviews with native and non-native Hewl\^eri individuals. The interviews are conducted based on a guide questionnaire. The questionnaire is adapted from \newcite{isam2024you}, which encourages participants to speak in conversation form and express themselves freely. For example, questions like \textit{"What is your earliest childhood memory?"} or \textit{"Can you tell what your typical day in your life?"} make participants think about personal stories and experiences, talking about their lives in their cultural context. The guideline avoids asking questions about private or sensitive topics in Hewl\^eri culture, such as family conflict, or marital status by concentrating on topics, such as daily routine and childhood memories. Examples of question topics are as follows:
\begin{itemize}[itemsep=2pt]
    \item Count numbers from 1 to 20.
    \item Personal background.
    \item Daily routine and experience.
    \item Hobbies and interests.
    \item The long responses (e.g., best film and favorite trip)
\end{itemize}

Participants are asked to sign a consent form by which they give permission to use the recorded voices for research purposes. Participants who are interviewed using online platforms (e.g.,  WhatsApp) sign the consent form digitally. We conduct pilot interviews to check and tune the suitability of the questionnaire and recording equipment. We use Audacity to record the conversations and export its .aup3 output to MP3 format to reduce the recording size. The records go through the following steps:

\begin{itemize}[itemsep=2pt]
    \item Cleaning background noise.
    \item Eliminating long spaces in the conversation.
    \item Removing any distractions that could draw away the participant attention.
    \item Stripping the interviewer's voice.
    \item Normalization to a consistent audio level.
    \item Exporting to WAV format.
\end{itemize}

We split the audio tracks into fixed-length segments using Pydub library. More specifically, we use \textit{AudioSegment} library to load the files and \textit{make\_chunks} library to produce fixed-track durations. The segmentation produces various-length segments, for example, three and five-second segments. \textit{Glob} library is used to locate and iterate over all relevant audio files.  MFCC (Mel Frequency Cepstral Coefficients) extracts the features, and the resulting data is used to train three models: ANN, RNN, and CNN. We uphold 10\% of the data for the final test. We also address data imbalance if necessary. The models are evaluated based on accuracy, training time, and human judgment.

\section{Results and Discussion}
This section presents the results of the research and discusses the findings.
\subsection{Data Collection}
\begin{figure}[H]
    \centering
    \includegraphics[width=0.60\linewidth]{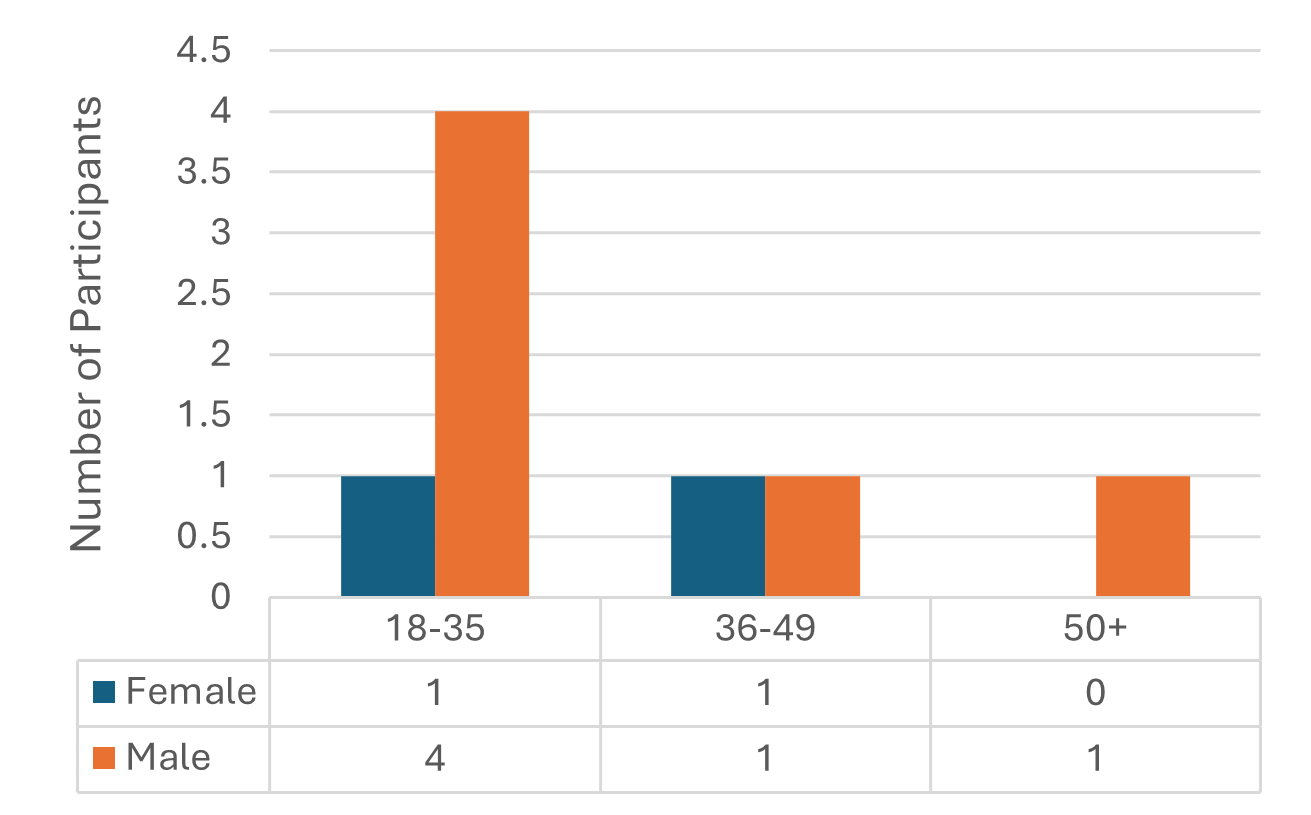}
    \caption{Gender and age distribution in the pilot study.}
    \label{fig:gender-pilot}
\end{figure}

We devised a questionnaire comprising five sections and 76 questions to guide the interviews. We conducted a pilot series of interviews to verify and adjust the questions, equipment, and outcome of the recordings. Eight individuals participated in the pilot interviews, which were conducted face-to-face. We identified a few issues related to the quality of the equipment and the recording process. All eight interviews were conducted face-to-face. We noticed that while some questions were assumed to receive different responses, they triggered similar ones. To encourage interviewees to respond that way, some other questions were revised to follow the Hewl\^eri subdialect (see Table \ref{11}). Also, based on the interviewee's response, we realized that it would be better to skip some questions. For example, “How old are your children?” should only be asked if the participant mentions that they have children. Consequently, comments and annotations were added to the questionnaire (see Table \ref{222}). Of this cohort, 75\% were native and 25\% non-native. Figure \ref{fig:gender-pilot} shows the gender and age distribution in the pilot study.

\begin{table}[H]
    \centering
    \caption{Questions that written in the Hewl\^eri subdialect}
    \begin{tabular}{|c|c|}
    
        \hline
        \rowcolor{lightgray}
        \textbf{Before piloting} & \textbf{After piloting} \\
        \hline
        \small \RL{هەولێری چاک و چۆنی کو دەکەن؟} & \small \RL{هەولێری کو سەلام لە یەکتری دەکەن؟} \\
        \hline
        \small \RL{ئەوەی تر چۆن جواب دەداتەوە؟} & \small \RL{ئەوەی دی کو جواب دەداتەوە؟} \\
        \hline
    \end{tabular}
    
    \label{11}
\end{table}

\begin{table}[H]
    \centering
    \caption{Questions modified with comments for skip logic }
    \begin{tabular}{|c|c|}
        \hline
        \rowcolor{lightgray}
        \textbf{Before Piloting} & \textbf{After Piloting} \\
        \hline
       \small \RL{مندارت هەیە؟ ناو و عومریان چەندە؟} & \small \RL{مندارت هەیە؟ ناو و عومریان چەندە؟ (لۆ خێزاندارەکان بە تەنێ)} \\
        \hline
         \small \RL{کوو بوو کەتیە ئەو قسمەی؟} & \small \RL{کوو بوو کەتیە ئەو قسمەی؟ (لۆئەوکەسانەی کە معهد یان کولیەیان خیندیە)} \\
        \hline
         \small \RL{سەعات چەند دەچیە دەوامێ؟} & \small \RL{سەعات چەند دەچیە دەوامێ؟ (لۆ ئەو کەسانەی کە دەوامیان هەیە)} \\
        \hline
        \small \RL{چ شیوەک دەزانی لێنێی؟} & \small \RL{چ شیوەک دەزانی لینێی؟ (زیاتر لۆ ئافرەتان)} \\
        \hline
    \end{tabular}
    
    \label{222}
\end{table}

The pilot interviews helped to consider other situations, for example, decreasing background noise by using an external microphone and recording the conversations in a reasonably quiet location. 

\subsection{Collected data}

Native participants were exclusively recruited from university students who were born in Hewl\^er to native Hewl\^eri-speaking parents and were continuously raised in Hewl\^er.  Non-native participants were selected from individuals who could speak the subdialect fluently but had different origins. To be qualified as non-native speakers, they should have continuously resided in the city for four years. Each participant completed a formal consent form to authorize the use of their voice recordings for academic research purposes. The Participation was voluntary; no incentives or compensations were offered.
 
The conversions were recorded using an HP EliteBook x360 1040 G7 laptop connected to a condenser microphone with a high-resolution specification of 192 KHz sampling rate at 24-bit depth. The microphone featured a low output resistance of 680 $\Omega$ with a frequency response range of 100-18000 Hz and a maximum sound pressure level (SPL) of 125 dB. Nearly 90\% of interviews were conducted face to face, while about 10\% were via communication platforms, such as Telegram, WhatsApp, or Viber.

We edited the records using Audacity version 3.7.1. All raw recordings were initially saved in .aup3 format. During the editing, interviewer questions were removed to retain only the responses, background noise was reduced, and clarity enhancement was applied to ensure a reasonable sound quality. The recordings were captured in stereo rather than mono, using a sample rate of 44,100 Hz and 16-bit signed PCM encoding to obtain a more natural and immersive auditory experience. Finally, the audio files were exported in WAV format using Pulse Code Modulation (PCM) encoding to maintain full audio fidelity for further linguistic or acoustic analysis. The total recording was 23 hours, 27 minutes, and 22 seconds. Table~\ref{tab:dataset_distribution} shows the details. 

\begin{table}[H]
\centering
\caption{Participant and duration distribution by speaker class.}
\begin{tabular}{|c|l|>{\centering\arraybackslash}p{4.2cm}|>{\centering\arraybackslash}p{3.6cm}|}
\hline
\rowcolor{lightgray}
\textbf{\#} & \textbf{Category} & \textbf{No. of participants} & \textbf{Total duration (HH:MM:SS)} \\ \hline
1 & Native & 19 (9 Female, 10 Male) & 10:51:32 \\ \hline
2 & Non-native & 21 (8 Female, 13 Male) & 12:35:50 \\ \hline

\multicolumn{2}{|l|}{Total} & {40} & {23:27:22} \\ \hline
\end{tabular}

\label{tab:dataset_distribution}
\end{table}

\begin{figure} [H]
    \centering
    \includegraphics[width=0.8\linewidth]{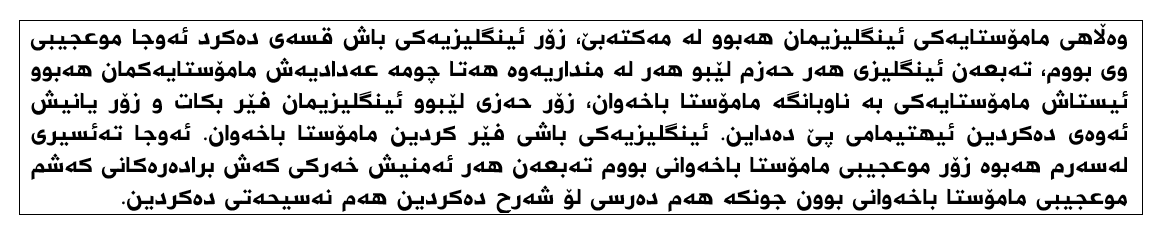}
    \caption{Native transcript (participant: Andam Muhsin, timeframe: 3:21-3:59).}
    \label{fig:native-transcript}
\end{figure}

\begin{figure} [H]
    \centering
    \includegraphics[width=0.8\linewidth]{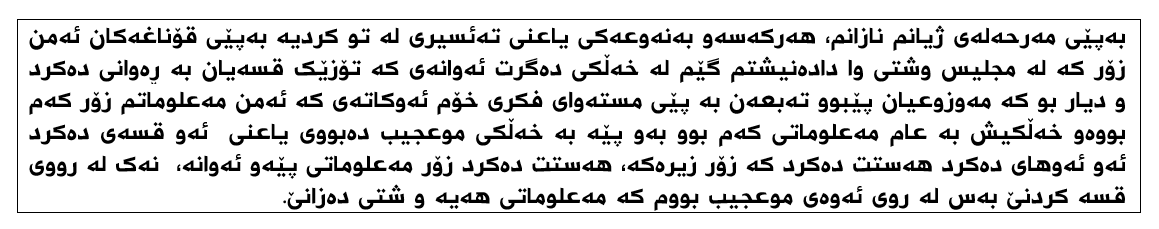}
    \caption{Non-native transcript (participant: Abdulqadir Omar, timeframe: 4:12-4:46).}
    \label{fig:Non_native_transcript}
\end{figure}

Of the total, 40.82\% of the participants were female, and 59.18\% were male. Of the total, 59.18\% were between 18 and 35, 28.57\% between 36 and 49, and 12.24\% were 50+ years old. Since dialects are dynamic and evolve over time, incorporating participants from multiple age groups allows the dataset to capture historical and generational features of the dialect, which in turn contributes to improved robustness and generalizability of our models. The majority of interviews (89.80\%) were conducted face-to-face, and 10.20\% were online. Table~\ref{tab:specs} provides a summary of the created dataset. Figures~\ref{fig:native-transcript} and~\ref{fig:Non_native_transcript} show two sample transcriptions from the dataset from a native and non-native speaker, respectively. 

\begin{table}[H]
\centering
\caption{Technical specifications of the Hewl\^eri speech corpus.}
\begin{tabular}{|l|l|}
\hline
\rowcolor{lightgray}
\textbf{Attribute} & \textbf{Specification} \\ \hline
Corpus name & Hewl\^eri Speech Corpus \\ \hline
Recording hardware & Condenser microphone \\ \hline
Recording software & Audacity \\ \hline
Total duration & 23 hours, 27 minutes, 22 seconds \\ \hline
Number of speakers & 40 \\ \hline
Average duration per speaker & 35.18 minutes (23h27m22s ÷ 40) \\  \hline
Frequency & 44,100 Hz \\ \hline
Sample rate & 22,050 Hz \\ \hline
File format & WAV \\
\hline
\end{tabular}

\label{tab:specs}
\end{table}

\subsection{Model development}
\begin{figure}[H]
    \centering
    \includegraphics[width=0.9\linewidth]{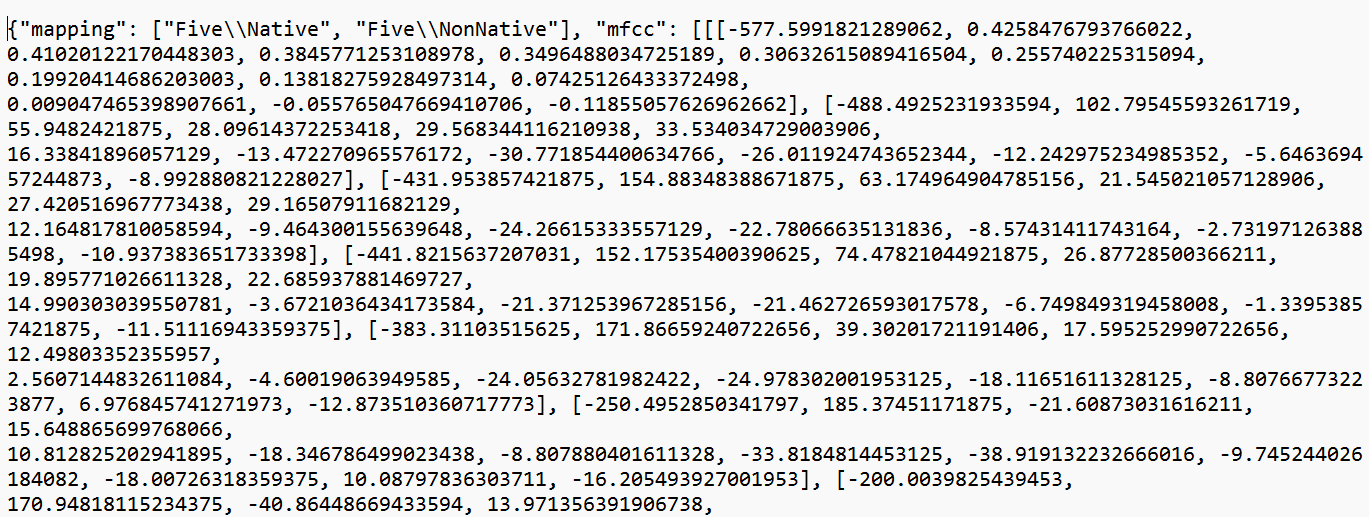}
    \caption{Json file content example}
    \label{fig:jsonfile}
\end{figure}

We used two environments to train the models: Google Colab and Jupyter Notebook. Google Colab was chosen because it provides a ready environment with no need for local setup. It enables seamless access to cloud datasets and offers high-speed GPU support, which significantly accelerates model training. On the other hand, Jupyter Notebook was used for preliminary experiments, data analysis, and local testing due to its interactive interface and flexible local execution. The following libraries were used for data preprocessing and model training: \texttt{numpy}, for numerical operations and array handling, \texttt{scikit-learn} (particularly, \texttt{train\_test\_split}) to divide data into training and testing sets, \texttt{tensorflow.keras}, to train ANN, CNN, and RNN models, and \texttt{matplotlib} to visualizing training performance and MFCC plots. We used \texttt{json} (Figure \ref{fig:jsonfile} shows a sample output) and \texttt{matplotlib} to visually inspect the extracted MFCC representations.

For voice segmentation and audio preprocessing we used \texttt{pydub} (specifically, \texttt{AudioSegment} and \texttt{make\_chunks}) to divide audio files into uniform chunks, and \texttt{glob} and \texttt{random} for file handling and sampling operations.

While most experiments were conducted on Google Colab, some were executed locally on a laptop with the following specifications: Intel Core i7-10610U CPU @ 1.80GHz, 16 GB of RAM, an 8 GB GPU, and Windows 11. These specifications were sufficient for smaller-scale experiments and model prototyping before full training was offloaded to cloud GPUs.

\begin{figure} [H]
    \centering
    \includegraphics[width=0.8\linewidth]{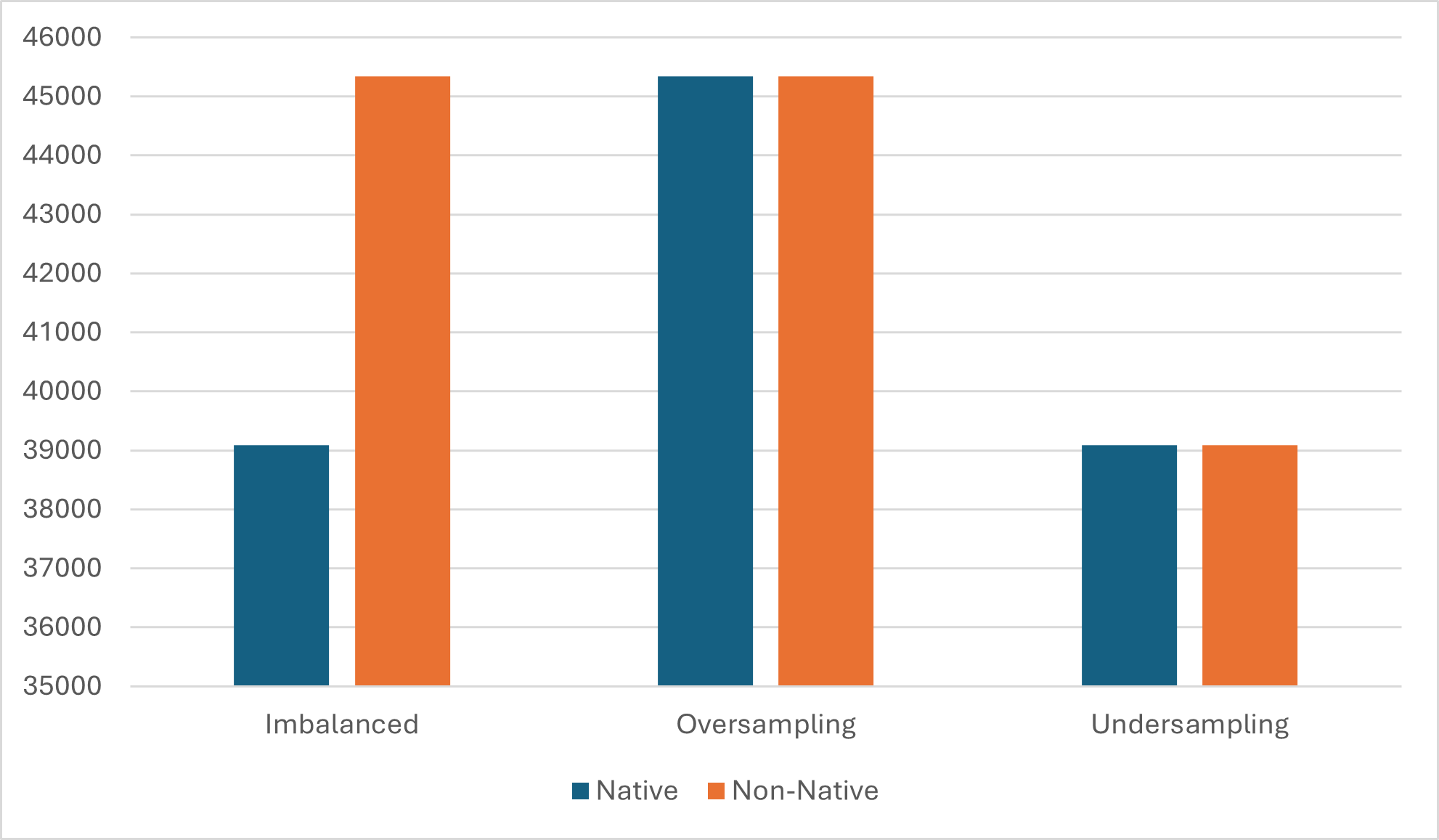}
    \caption{Class distribution before/after balancing: oversampling vs. undersampling }
    \label{fig:Imbalance_over_under}
\end{figure}

As mentioned, the collected data was imbalanced in terms of its two classes. We used random oversampling to increase the minority class data by duplicating its samples to make it equal to the majority class samples (see Figure ~\ref{fig:Imbalance_over_under}). That increased the native samples from 39083 to. Similarly, we used random sampling to decrease the non-native sample size from 45340 to 39083.

\subsection{Experiments}
We conducted 66 comprehensive experiments on three different models (ANN, CNN, RNN) to analyze the impact of various data splitting schemes and over- and undersampling on the models' performance. We used seven different dataset truck durations: one-second, three-second, five-second, 10-second, 20-second, 30-second, and 60-second segmenting schemes. For all cases, 10\% of the data was held for the final evaluation. Of the 90\% of the data, the splitting scheme for ANN was 80:20 for the training and testing sets, and for CNN and RNN, 80:10:10 for training, validation, and testing.
Early stopping was applied to prevent overfitting with a patience threshold of 10 epochs. The models' performance was measured using cross-entropy loss and classification accuracy, which was recorded at the best early stopping triggered.

ANN achieved its highest accuracy at 82.92\% on the oversampled dataset with 5-second segments of audio truck duration, while its accuracy was 46.98\% in 60-second segments (see Appendix A, Figures~\ref{fig:ann_Exp} and~\ref{fig:ann} for more details). That was, as we understood, because shorter audio truck duration could better preserve time-dependent features that uniquely identified a class, which has been reported by other researchers (see \newcite{1021072}). Applying k-fold cross-validation, the accuracy was 82.11\% on held-out data. Although this result showed a 0.81\% decrease in the model performance, we could still observe a strong generalizability of the model.

RNN achieved its highest accuracy at 95.92\% on the oversampled dataset with 5-second segments of audio truck duration, while its accuracy was 70.54\% in 60-second segments (see Appendix A,  Figures~\ref{fig:rnn-exp} and \ref{fig:rnn} for more details).  

CNN achieved its highest accuracy at 94.47\% on the oversampled dataset with 10-second segments of audio truck duration, while its accuracy was 79.65\% in 60-second segments (see Appernid A, Figures~\ref{fig:cnn-exp} and \ref{fig:cnn} for more details). 

\subsection{Models comparison}
RNN performed better, comparing with ANN and CNN, due to its sequential modeling skills, which captured evolving dialectal features over short periods of time (3-10 seconds). For balanced or undersampled datasets, the accuracy for audio segments more than 20 seconds dropped by~15\%. CNN model closely followed at 94.47\% peak accuracy, demonstrating its robustness with imbalanced data and different segment lengths. Its performance was resilient due to the nature of its convolutional operations, which enabled consistent performance across various configurations even on 60-second undersampled data.

From the computation perspective, RNN with GPU acceleration was trained 3x faster than CPU-based ANN/CNN. CNN provided the ideal middle ground, achieving over 90\% accuracy in 75\% of experiments conducted solely on CPU. ANN achieved a moderate peak accuracy of 82.92\%, but unmatched generalization capabilities only dropping by 0.81\%, making it the most robust choice for deployment scenarios characterized by high variability within data streams. RNN model dropped by 0.69\% after applying K-fold cross-validation generalization methods, and dropped 1.05\% with hold-out data. Similarly, for CNN there are not much differences (2.34\%, 2.32\%) after applying generalization techniques.

Various trends were apparent in all models; performance decreased as duration increased, and the highest performance was cited between one and ten seconds; oversampling provided a 7–12\% increase in accuracy compared to undersampling and was the most effective data balancing strategy. RNNs effectively captured temporal dynamics, while CNN recognized structure at more complex levels. ANN showed better behavior for generalization.

\subsubsection{Comparison of human and machine prediction}

We conducted a single-track evaluation experiment, in which RNN, trained on oversampled data with a 5-second track duration, outperformed ANN and CNN. 

Two separate 5-second soundtracks were prepared, one from a native Hewl\^eri speaker and another from a non native one. These samples were passed to the trained RNN model for classification. The same tracks were also given to humans to classify. for further testing the model the same RNN model who were trained on 5 second sound track was evaluated on 60 second (one minute) samples, and then we have reversed the process another RNN model trained on the 60 second sound track was evaluated by using the same both 5 second and 60 second samples as it shown in the Table \ref{tab:single_track_evaluation}.

\begin{table}[H]
\centering
\caption{Comparison of human and model predictions in single track evaluation}
\begin{tabular}{| p{2.5cm} | p{2.5cm} | p{3cm} | p{3cm} | p{3cm}|}
\hline
\rowcolor{lightgray}
\textbf{Duration} & \textbf{Actual} & \textbf{Human} & \textbf{Model (5s)} & \textbf{Model  (60s)} \\
\hline
5 seconds   & Native       & Native/Non-Native & Native      & Native      \\
\hline
5 seconds   & Non-Native   & Non-Native           & Non-Native  & Non-Native  \\
\hline
60 seconds  & Native       & Native               & Non-Native  & Native      \\
\hline
60 seconds  & Non-Native   & Non-Native           & Non-Native  & Non-Native  \\
\hline
\end{tabular}

\label{tab:single_track_evaluation}
\end{table}

Most of the time, both computer model and human predictions were on a par with the actual nativity of the speaker, particularly when the soundtrack durations were the same as the training duration. Furthermore, we observed that the model's performance decreases when the duration of the input sample is longer. In other words, human listeners benefit from the longer speech samples, as it allows them to decide based on more information. However, that delayed decision could be problematic in some tasks that apply the NLI. 

\subsubsection{Computational costs}
All three models were trained and implemented using Google Colab. The ANN and CNN were trained using CPU runtime type, while the RNN was trained on T4 GPU run time type due to requiring a lot of sequential computation. The CNN required the longest training time (3,095 seconds ), ANN, 1,675 seconds, and RNN, 453 seconds (see Table~\ref{tab:training-time}).

\begin{table}[H]
\centering
\caption{Comparison of models' computational cost}
\begin{tabular}{|c|c|c|c|}
\hline
\rowcolor{lightgray} 
\textbf{Model} & \textbf{Best accuracy} & \textbf{Training time (seconds)} & \textbf{Hardware used} \\
\hline
ANN & 82.92\% & 1675.688 & CPU \\
\hline
CNN & 94.47\% & 3095.492 & CPU \\
\hline
RNN & 95.92\% & 453.051 & GPU (T4) \\
\hline
\end{tabular}

\label{tab:training-time}
\end{table}

 The RNN model achieved the highest performance with the shortest time. CNN achieved strong accuracy but was computationally expensive, while ANN provided moderate balanced, and efficient performance between accuracy and computational cost. This comparison shows a clear relationship between hardware configuration and efficiency of the model. The training time of CNN is more than doubles of the ANN and is nearly seven times longed than that of the RNN. This indicates that, along with the capability of CNN to capture complex features it requires considerable computational cost when trained on a CPU in Google Colab.

\subsubsection{Comparison with best practices}
We compared our results with three widely accepted standard models in the field of NLI. The first benchmark is the model by \newcite{malmasi2017report} in 'A report on the Native Language Identification Shared Task 2017', which used an SVM (Support Vector Machine) that was trained on essays and spoken responses from test 
takers of the TOEFL iBT exam. The model achieved 93.19\% while combining text and speech.
The second reference is the work by \newcite{malmasi2017multilingual}, which used an SVM across six non Native English learners dataset. It achieved 95\% accuracy in distinguishing between native and non native speakers. The third benchmark is the research by \newcite{brooke2012robust} that solved the issue of topic bias using cross-corpus evaluation and bias-adapted SVMs. The accuracy was 95\% using their lexical system, especially using Asian language groups and using large learner corpora such as Lang-8 and ICLE. Table \ref{tab:comparison_nli} shows the comparison results.

\begin{table}[H]
\centering
\caption{Comparison of benchmark NLI models and the proposed Hewl\^eri NLI}
\renewcommand{\arraystretch}{1.2}
\setlength{\tabcolsep}{5pt}

\begin{tabular}{|p{3.2cm}|p{4.6cm}|p{3.2cm}|p{2.2cm}|}
\hline
\rowcolor{lightgray}
\textbf{Study} & \textbf{Methodology} & \textbf{Dataset} & \textbf{Accuracy (\%)} \\ \hline

Malmasi et al. (2017) \newline \textit{NLI Shared Task 2017} & 
SVM trained on essays and spoken TOEFL iBT responses (text + speech) & 
TOEFL iBT (multiple L1s) & 
93.19 \\ \hline

Malmasi \& Dras (2017) \newline \textit{Multilingual NLI} & 
SVM applied to six non-native English learner datasets & 
Multilingual (non-native English speakers) & 
95.00 \\ \hline

Brooke \& Hirst (2012) \newline \textit{Robust Lexicalized NLI} & 
Lexicalized SVM using cross-corpus evaluation and topic-bias reduction & 
Lang-8, ICLE (Asian language groups) & 
95.00 \\ \hline

{Proposed Hewl\^eri NLI} & 
Deep learning (RNN with oversampling; 5-sec audio chunks) & 
Kurdish (Hewl\^eri subdialect) & 
{95.92} \\ \hline

\end{tabular}
\label{tab:comparison_nli}
\end{table}

\section{Conclusion}
We collected 23 hours and 27 minutes of speech, consisting of native and non-native participants in Hewl\^eri sudialect: 10 hours and 51 minutes of native and 12 hours and 36 minutes of non-native. The data were segmented into multiple short samples with durations of one, three, five, 10, 20, 30, and 60 seconds. We used MFCC for feature extraction and trained three models: ANN, CNN, and RNN. The RNN model achieved the highest accuracy at 95.92\%, the CNN 94.47\%, and the ANN 82.99 while providing a robust baseline across all segment durations.

\section*{Ethical Consideration}
The participants provided their consent to use and publicize the recordings. The publicly released dataset associated with this study has been processed to remove any sensitive personal information that could be traced back to any particular individual to minimize potential risks to individuals whose tweets were included.

\section*{Author Contributions}
Conceptualization, Hardi Garari (H.G.) and Hossein Hassani (H.H.); methodology, H.G.; software, H.G.; validation, H.G. and H.H.; formal analysis, H.G.; investigation, H.G.; resources, H.G. and H.H.; data curation, H.G.; preparing first draft, H.G.; revising the draft and preparing final manuscript H.H.; visualization, H.G.; supervision, H.H.; project administration, H.H.

\section*{Dataset Availability}
The dataset will be publicized on KurdishBLARK but it may take some time before it becomes accessible. 

\section*{Acknowledgment}
The authors extend their gratitude to all participants in the data collection process. The names of those who agreed to be publicly mentioned appear in Appendix B. 

\bibliographystyle{lrec}
\bibliography{sample}

\newpage
\appendix
\setcounter{page}{1}
\renewcommand{\thepage}{A\arabic{page}}
\section*{Appendix A - Experiment Results}

\begin{figure}[H]
    \centering
    \includegraphics[width=0.7\linewidth]{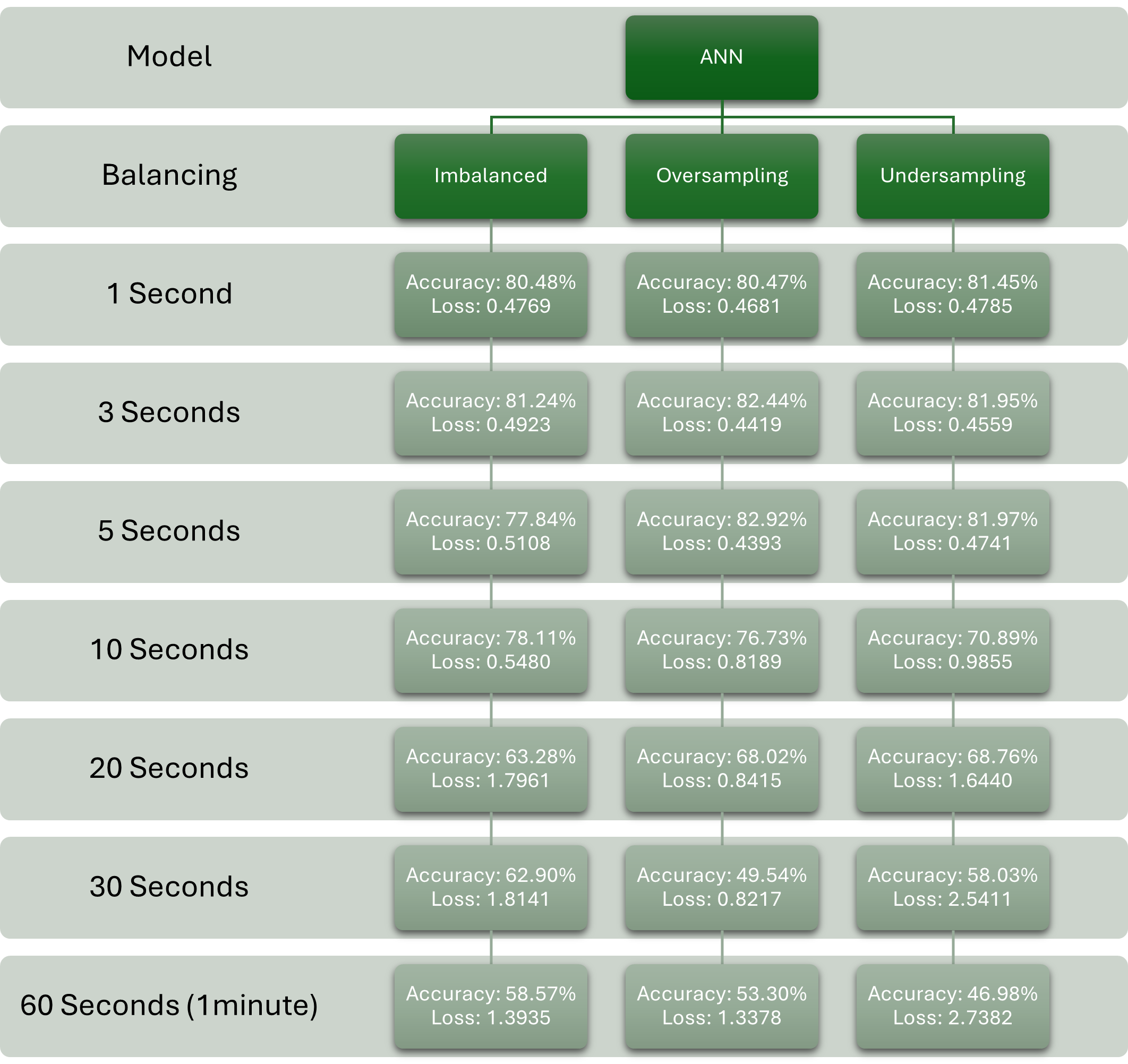}
    \caption{ANN Experiment's result (accuracy and loss)}
    \label{fig:ann_Exp}
\end{figure}

\begin{figure}[H]
    \centering
    \includegraphics[width=0.7\linewidth]{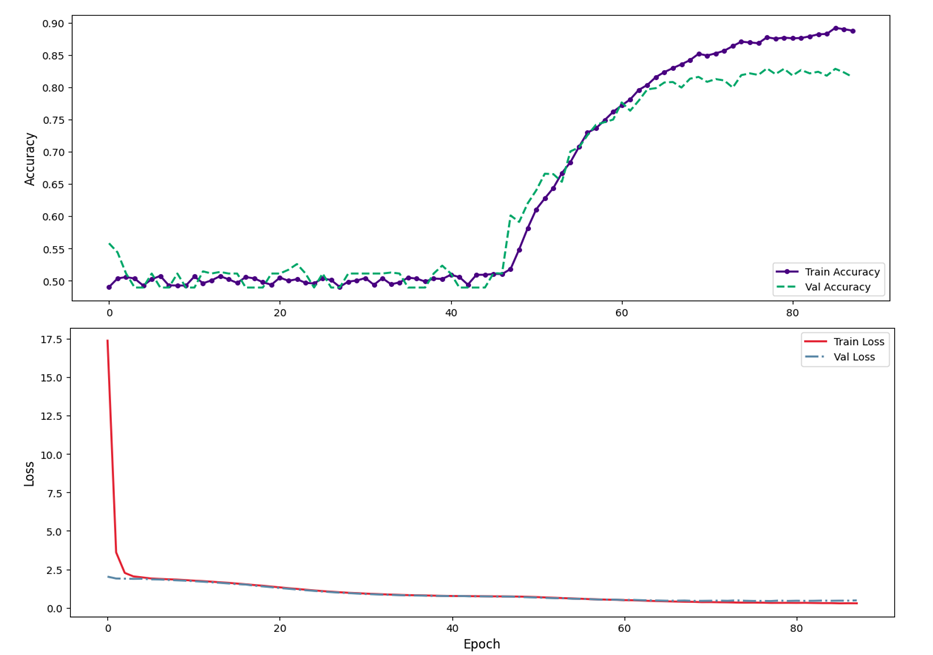}
    \caption{ANN model accuracy and loss over epochs}
    \label{fig:ann}
\end{figure}

\begin{figure}[H]
    \centering
    \includegraphics[width=0.7\linewidth]{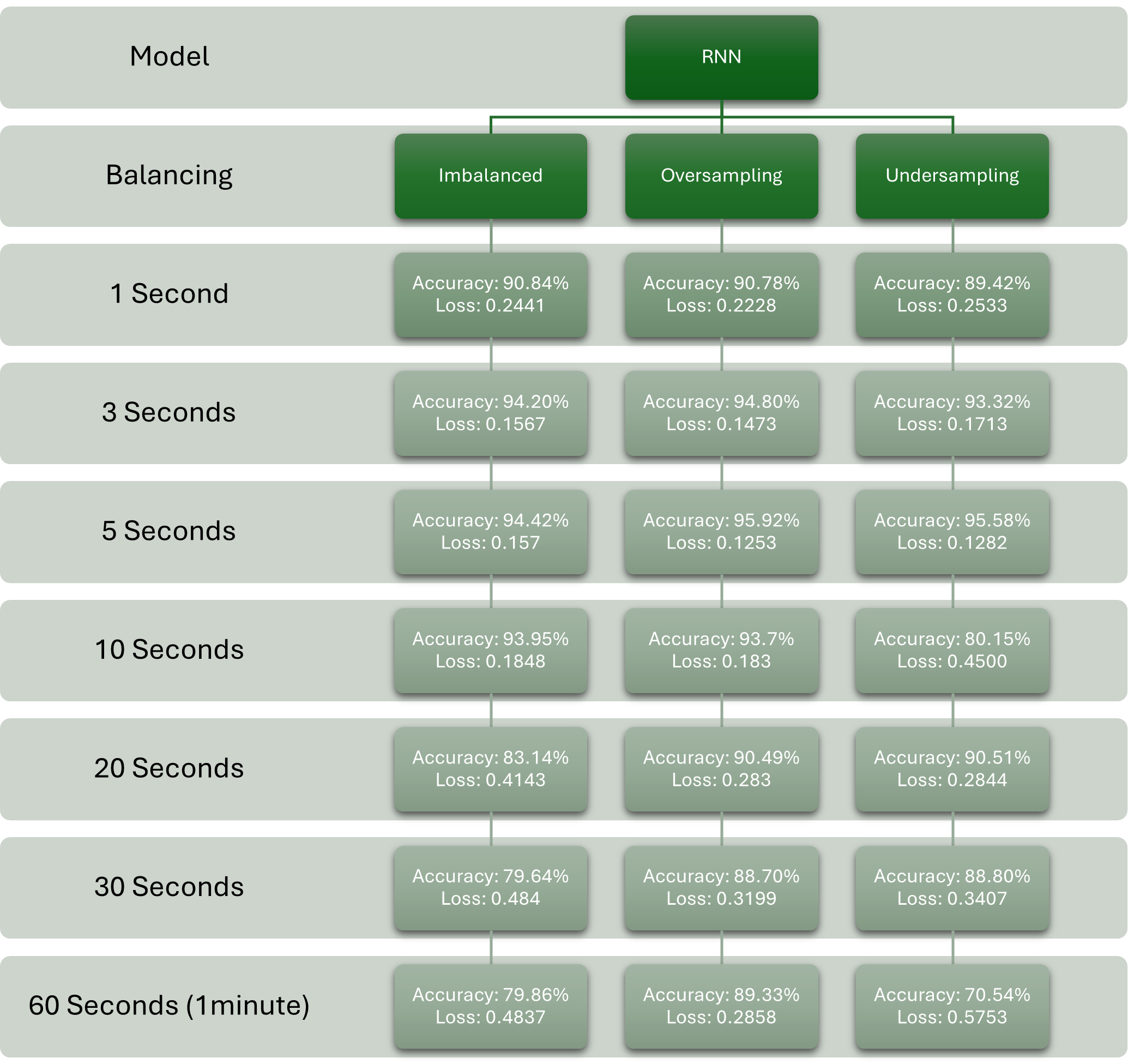}
    \caption{RNN experiment's result (accuracy and loss)}
    \label{fig:rnn-exp}
\end{figure}

\begin{figure}[H]
    \centering
    \includegraphics[width=0.7\linewidth]{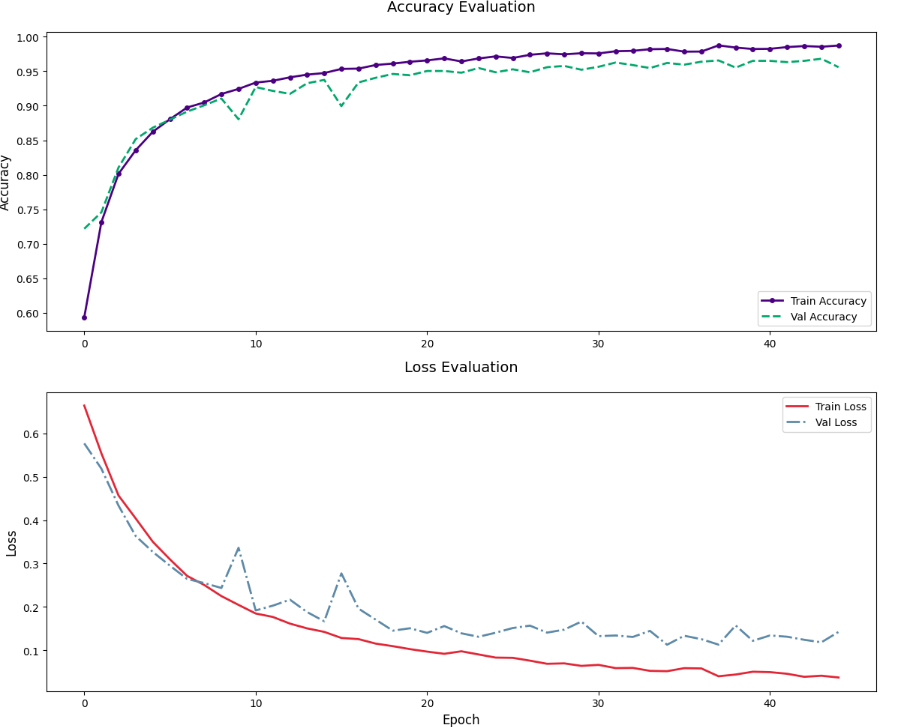}
    \caption{RNN model accuracy and loss over epochs}
    \label{fig:rnn}
\end{figure}

\begin{figure}[H]
    \centering
    \includegraphics[width=0.7\linewidth]{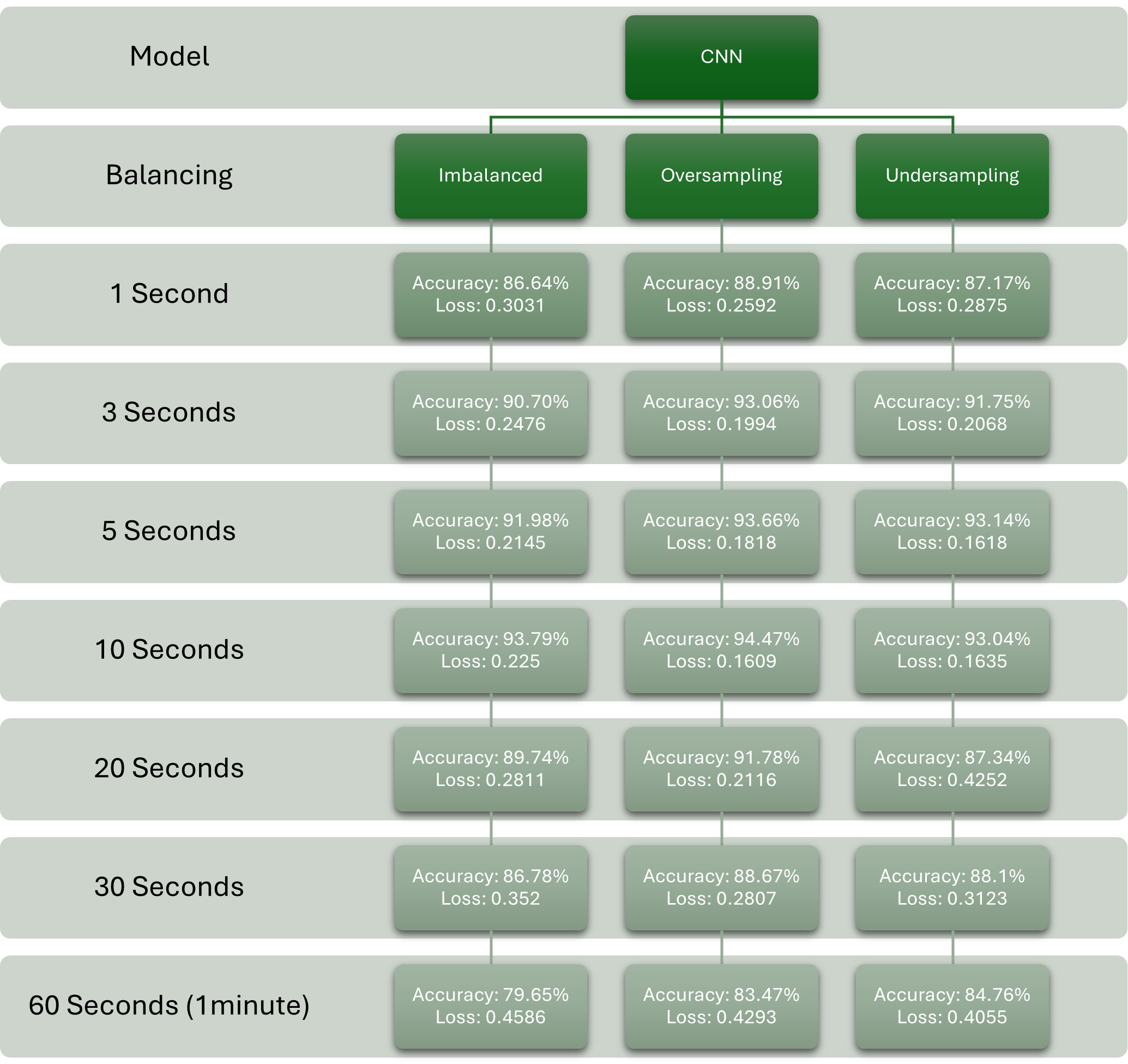}
    \caption{CNN experiment's result (accuracy and loss)}
    \label{fig:cnn-exp}
\end{figure}
\begin{figure}[H]
    \centering
    \includegraphics[width=0.7\linewidth]{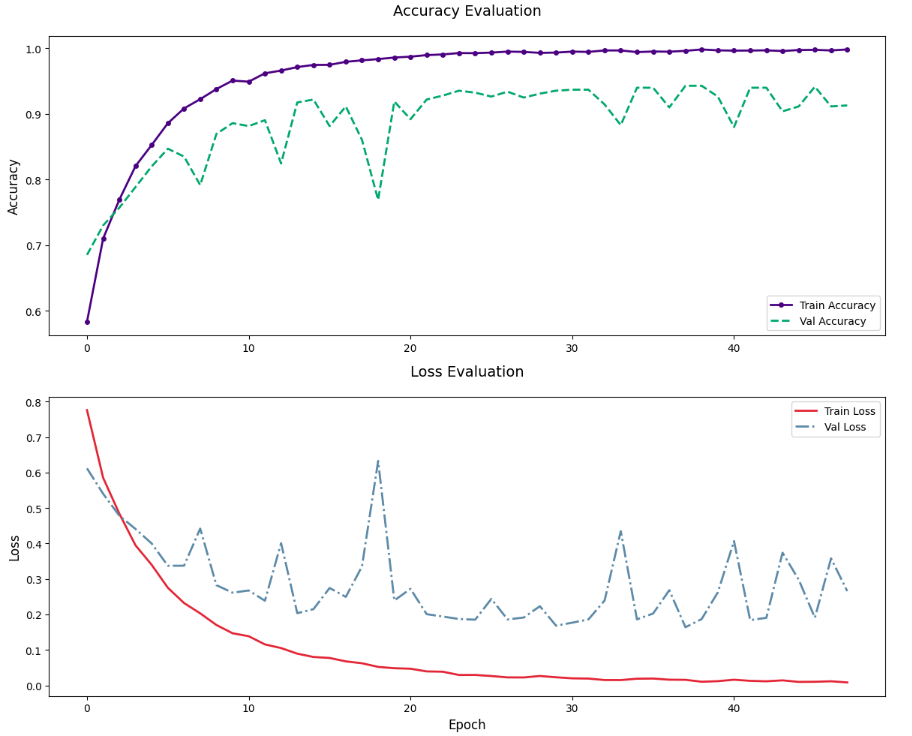}
    \caption{CNN model accuracy and loss over epochs}
    \label{fig:cnn}
\end{figure}

\newpage
\appendix
\setcounter{page}{1}
\renewcommand{\thepage}{B\arabic{page}}
\section*{Appendix B - Name of participants}

The following list includes the names of participants in both pilot and main data collection interviews. The list excludes those who wished to remain anonymous.

\setlength{\parindent}{0cm}
\begin{multicols}{2}
\begin{itemize}[itemsep=0.1cm]

\item Abdalla Muhammed Salih
\item Abdulqadir Omar Faqe
\item Ahmed Ali
\item Ahmed Mahmood Abdulrahman
\item Andam Muhsin 
\item Arsalan Wahid Hassan
\item Awsama Kamal Omar
\item Bala Kawa
\item Bnar Jawhar Aziz
\item Bzhar Dlshad Hussien
\item Daban Othman Muhammed
\item Diyar Tahir Aziz
\item Eman Wahid Khalid
\item Faisal Muhammed ShafaHassan
\item Fatima Ismail Qadr
\item Hazhar Qadr Ismail
\item Helin Sardar Azeez
\item Hero Sardar Azeez
\item Jinan Mukaram Ahmed
\item Kabir Mirkhan Ahmed
\item Khumar Jawhar Aziz
\item Khunaw Jawhar Aziz
\item Lana Rebar
\item Muhammed Abdulla Younes
\item Muhammed Hunar Mahmood
\item Muhammed Mahdi Salih
\item Mustafa Faris Fadhel
\item Nashmeel Hamad Sadq
\item Nsar Jawhar Aziz
\item Ramazan Sleman Hamk
\item Rangin Sabah Hassan
\item Rawand Azad
\item Renas Adnan Omar
\item Sami Raqeeb Ahmed
\item Sana Isam Taib
\item Sazan Bayar Osman
\item Shawnm Rzgar
\item Shler
\item Vian Wahid Khalid
\item Zana Himdad 
\item Zhyar Muhammed Amin Ahmed
\item Zmnako Bahaddin Taha
\end{itemize}
\end{multicols}

\end{document}